\documentclass[10pt,twocolumn,letterpaper]{article}

\usepackage{iccv}
\usepackage{times}
\usepackage{epsfig}
\usepackage{graphicx}
\usepackage{amsmath}
\usepackage{amssymb}
\usepackage{xcolor}
\usepackage{mathrsfs}
\usepackage{booktabs}
\usepackage{comment}
\usepackage{multirow}
\usepackage{url}
\usepackage{dblfloatfix}
\usepackage[title]{appendix}
\usepackage{placeins}

\newcommand{\norm}[1]{\lVert #1 \rVert}

\newcommand{\model}{SplitNet\xspace}

\newcommand{\thirddataset}{IndoorEnv}

\usepackage[breaklinks=true,bookmarks=false]{hyperref}

\iccvfinalcopy 

\def\iccvPaperID{3164} 

\ificcvfinal\fi

\begin{document}

\title{\model{}: Sim2Sim and Task2Task Transfer \\ for Embodied Visual Navigation}

\author{Daniel Gordon$^1$\thanks{Work done during an internship at Facebook AI Research.} \quad Abhishek Kadian$^2$ \quad Devi Parikh$^{2,3}$ \quad Judy Hoffman$^{2,3}$ \quad Dhruv Batra$^{2,3}$ \\
{\normalsize $^1$Paul G. Allen School of Computer Science, University of Washington} \\
{\normalsize $^2$Facebook AI Research} \ 
{\normalsize $^3$Georgia Institute of Technology}
}
\maketitle
\ificcvfinal\thispagestyle{empty}\fi

\begin{abstract}
We propose \model{}, a method for decoupling visual perception and policy learning.
By incorporating auxiliary tasks and selective learning of portions of the model, we explicitly decompose the learning objectives for visual navigation into perceiving the world and acting on that perception. We show improvements over baseline models on transferring \textbf{between simulators,} an encouraging step towards Sim2Real. Additionally, \model{} generalizes better to unseen environments from the same simulator and transfers faster and more effectively to novel embodied navigation tasks. Further, given only a small sample from a target domain, \model can match the performance of traditional end-to-end pipelines which receive the entire dataset
\footnote{\url{https://github.com/facebookresearch/splitnet}}

\end{abstract}



\vspace{-5mm}
\section{Introduction}
A longstanding goal of computer vision is to enable robots to understand their surroundings, navigate efficiently and safely, and perform a large variety of tasks in complex environments. A practical application of the recent successes in Deep Reinforcement Learning is to train robots with minimal supervision to perform these tasks. Yet poorly-trained agents can easily injure themselves, the environment, or others. These concerns, as well as the difficulty in parallelizing and reproducing experiments at a low cost, have drawn research interest towards simulation environments~\cite{mp3d, ai2thor, gibson, chalet, habitat19arxiv}.

\begin{figure}[t]
\begin{center}
\includegraphics[width=1.0\columnwidth]{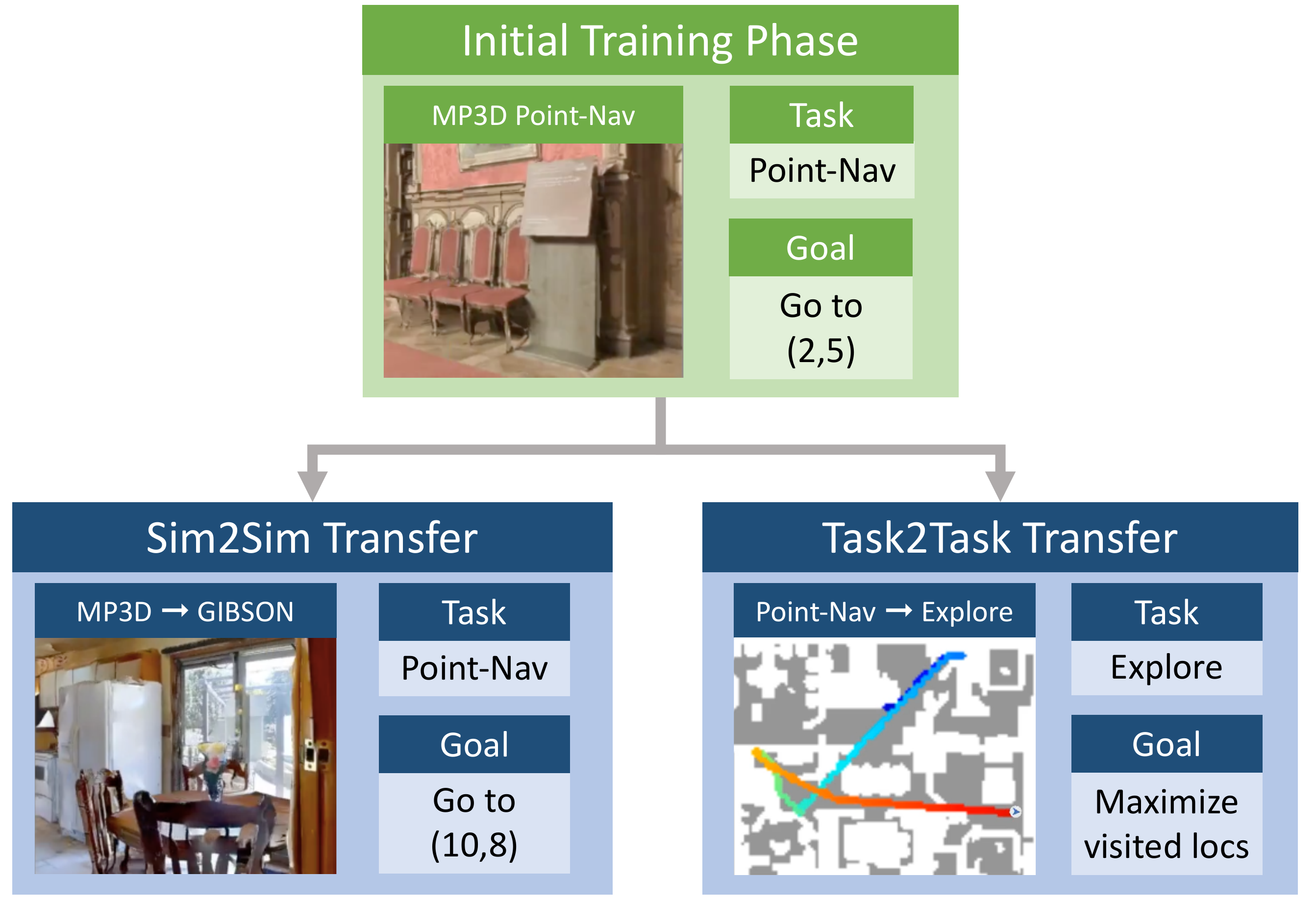}
\caption{We decompose learning of visual navigation tasks into learning of a visual encoder and learning of an embodied task decoder. Through this decomposition we enable fast transfer to new visual environments and transfer to new embodied tasks. 
}
\vspace{-8mm}
\label{fig:teaser}
\end{center}
\end{figure}

However no simulator perfectly replicates reality, and agents trained in simulation often fail to generalize to the real-world. Transferring learned policies from simulation to the real-world (Sim2Real) has become an area of broad interest~\cite{peng2018sim, rusu2017sim, tobin2017domain} yet there still exists a sizable performance gap for most algorithms. Furthermore, Sim2Real transfer reintroduces safety and reproducibility concerns. To mitigate this, we explore the related task of Sim2Sim, transferring policies between simulators, for embodied visual navigation (Figure~\ref{fig:teaser}). Transferring between simulators incurs a similar ``reality gap'' as between simulation and reality, due to differences in data collection and rendering. Learning to transfer between simulation environments serves as an encouraging preliminary step towards true Sim2Real transfer.

To enable Sim2Sim transfer we propose \model{}, a composable model for embodied visual tasks which allows for the sharing and reuse of information between different visual environments. \model{} enables transfer across different embodied tasks (Task2Task), meaning our model can learn new skills quickly and adapt to the ever-changing requirements of end users. Our key insight is to observe that embodied visual tasks are naturally decomposable into visual representation learning to extract task agnostic salient information from the visual input, and policy learning to interpret the visual representation and determine a proper action for the agent. Rather than learning these components solely independently or completely tied, we introduce an algorithm for learning these embodied visual tasks which benefits both from the scalability and strong in-domain, on-task performance of an end-to-end system and from the generalization and fast adaptability of modular systems. 

\model{} incorporates auxiliary visual tasks, such as depth prediction, as a source of intermediate supervision which guides the visual representation learning to extract information from the images extending beyond the initial embodied task. We demonstrate that initial pre-training of the visual representation on such auxiliary visual tasks produces a more robust initialization than the standard approach of pre-training on auxiliary visual datasets (e.g. ImageNet~\cite{imagenet}) which may not be from an embodied perspective. Then we showcase the composability of our model by illustrating its ability to selectively adapt only the visual representation (when moving to a new visual environment) or only its policy (when moving to a new embodied task). 

We center our evaluation on adapting between different simulators of varying fidelity and between different embodied tasks. Specifically, our experiments show that compared to end-to-end methods, \model{} learns more transferable visual features for the task of visual point-to-point navigation, reduces overfitting to small samples from a new target simulator, and adapts faster and better to novel embodied tasks. 

In summary, our contributions are \textbf{1.} a principled way to decouple perception and policy (in our case navigation), \textbf{2.} showing that this technique improves performance on the primary task \textit{and} facilitates transfer to new environments and tasks, \textbf{3.} describing specifically how to update the weights for new tasks or new environments using significantly less data to adapt.
\section{Related Work}

This work introduces a learning approach for transferring visual representations between environments and for transferring policy information between different embodied tasks. The most related lines of work focus on adaptation and transfer of visual representations, deep reinforcement learning (especially from visual inputs), and transferring from simulation to the real-world (Sim2Real) both for visual and embodied tasks. 

\vspace{-3mm}
\paragraph{Visual Transfer and Adaptation.}
Many works have explicitly studied techniques for increasing the reusability of learned information across different visual tasks. Domain adaptation research has mainly focused on reusing a representation even as the input distribution changes, with most work focusing on representation alignment through explicit statistics~\cite{2015icml_long,2016eccv_sun} or through implicit discrepancy minimization with a domain adversarial loss~\cite{2015icml_ganin,2017cvpr_tzeng}. A related line of work focuses on sharing between two image collections through direct image-to-image transfer~\cite{dcgan,cyclegan}, whereby a mapping function is learned to take an image from one domain and translate it to mimic an image from the second domain~\cite{2017cvpr_bousmalis,2018icml_hoffman,cogan,2017iclr_taigman}.

In parallel, many works focus on reusing learned representations for solving related visual tasks. The most prevalent such technique is simply using the first representation parameters as initialization for learning the second, termed finetuning~\cite{2014cvpr_rcnn}. A recent study proposed a technique for computing the similarity between a suite of visual tasks to create a Taskonomy~\cite{taskonomy2018} which may be used to determine, given a new task, which prior tasks should be used for the initialization before continued learning. This method focuses on ``passive'' visual understanding tasks such as recognition, reconstruction and depth estimation and does not delve into learning representations for ``active'' tasks such as embodied navigation where an agent must both understand the world and directly use its understanding for some underlying task.

Overall, much of the prior work has focused on representation learning for visual recognition. In contrast, this work studies transfer of visuomotor policies for embodied tasks and decomposes the problem into transfer of visual representations for embodied imagery (Sim2Sim) and transfer of policies across various downstream embodied tasks (Task2Task). 

\vspace{-3mm}
\paragraph{Visual RL Tasks:}
In parallel with the development of deep representation learning for passive visual tasks, there has also been a plethora of recent research on policy learning from visual inputs inspired by the success of end-to-end visuomotor policy learning~\cite{2016jmlr_levine, levine2018learning, mnih2016asynchronous}. Much of the success here comes from training on large-scale~\cite{levine2018learning} data, frequently made possible by extensive use of simulation environments~\cite{gupta2017cognitive, mnih2016asynchronous, curiosity, 2017icra_zhu}. These techniques often leverage the additional supervision and auxiliary tasks given by the simulators to bootstrap their learning~\cite{2017iclr_mirowski, perceptualactor}. Perceptual Actor~\cite{perceptualactor} specifically examines how 20 different pretraining tasks affect the learning speed and accuracy of a visual navigation policy as compared to random initialization. Others use unsupervised~\cite{jaderberg2016reinforcement} or self-supervised~\cite{curiosity} learning as an additional signal in domains with sparse rewards. We build on these approaches by explicitly separating the auxiliary learning from the policy layers to ensure a decoupling of the weights which enables better transfer to new environments. 

For increased task generalization, others have proposed using the successor representation~\cite{2017iros_zhang, zhu2017visual} which decomposes the reward and Q-functions into a state-action feature $\phi_{s,a}$, a successor feature $\psi_{s,a}$ and a task reward vector $w$. This decomposes the network into one which learns the dynamics of the environment separate from the specified task, which allows for faster transfer to new tasks by only retraining the task embedding $w$. Our proposed method allows quick transfer to new tasks as well as new environments.

\vspace{-3mm}
\paragraph{Sim2Real:}
Significant progress has been made on adapting between simulated and real imagery for visual recognition, especially in the context of semantic segmentation in driving scenes~\cite{2018icml_hoffman,fcninwild,gam,curriculumDA}. These techniques build on the visual domain adaptation methods described above. In parallel, there has been work on transferring visual policies learned in simulation to the real-world, but often limited to simple visual domains~\cite{rusu2017sim, tobin2017domain} which bear little resemblance to the complexity of true real-world scenes. Rusu et al.~\cite{rusu2017sim} train a network in simulation before initializing a new network which receives outputs from the simulation-trained network as well as real-world inputs. Yet their evaluation is limited to simple block picking experiments with no complex visual scenes. Peng et al.~\cite{peng2018sim} use randomization over the robot dynamics to learn robust policies, but do not use visual inputs in simulation or reality and only perform simple puck-pushing tasks. Sadeghi et al.~\cite{sadeghi2017cadrl} also uses randomization of textures, lighting, and furniture placement in a simulation environment for Sim2Real transfer of drone flight. Tobin et al.~\cite{tobin2017domain} and Sadeghi et al.~\cite{sim2real_viewpoint} randomize colors, textures, lighting, and camera pose as a form of augmentation of the simulated imagery to better generalize to real-world imagery for picking tasks. \cite{tobin2017domain} focuses on primitive geometric objects for picking tasks and does not decouple visual feature learning from policy learning which limits the transferrability of their method to new tasks. \cite{sim2real_viewpoint} shows similar benefits of decoupling perception and policy for Sim2Real transfer, but do not explore transfer to new tasks. Two recent method~\cite{2018eccv_drivingsim2real, divis}, uses semantic segmentation and obstacle detection as an intermediate objective to aid in transferring learned  driving policies from simulation to the real-world. While we do not transfer our policies to real robots, we focus on visually diverse scenes which better match the complexity of the real-world then the simplistic setups of many of the prior policy transfer approaches. Similar to M{\"u}ller et al.~\cite{2018eccv_drivingsim2real} we use auxiliary intermediate objectives to aid in transfer, but in our case focus on a set of auxiliary visual and motion tasks which generalize to many downstream embodied tasks and propose techniques to selectively transfer either across visual environments or across embodied tasks.

\section{\model: Decoupled Perception and Policy}

\newcommand{\recon}{\mathcal{R}}
\newcommand{\predNext}{\mathcal{P}}
\newcommand{\ego}{\mathcal{E}}
\newcommand{\surfnorm}{\mathcal{S}}
\newcommand{\depth}{\mathcal{D}}
\newcommand{\visft}{\phi}
\newcommand{\image}{I}
\newcommand{\action}{a}
\newcommand{\encoder}{\mathcal{F}}
\newcommand{\decoder}{\mathcal{G}}


Solving complex visual planning problems frequently requires different types of abstract understanding and reasoning based on the visual inputs. In order to learn compact representations and generalizable policies, it is often necessary to go beyond the end-to-end training paradigm. This is especially true when the initial learning setting (source domain) and current learning setting (target domain) have sufficiently different visual properties (e.g. differing visual fidelity as seen in Figure \ref{fig:teaser} \textit{left}) or different objectives (e.g. transfer from one task to another as in Figure~\ref{fig:teaser} \textit{right}). In this section we outline the learning tasks we use, and our strategy for training a network which transfers to new visual domains and new embodied tasks.

\subsection{Embodied Tasks}
\label{sec:tasks}
In this work, we focus on the following three visual navigation tasks which require memory, planning, and geometric understanding: Point-to-Point Navigation (Point-Nav), Scene Exploration (Exploration), and Run Away from Location (Flee). In our experiments, all tasks share a discrete action space: Move Forward by 0.25 meters and Rotate Left/Right by 10 degrees.

\textbf{Point-to-Point Navigation (Point-Nav)}
\label{sec:pointnav}
An agent is directed to go to a point via a constantly updating tuple of (angle to goal, distance to goal). The agent succeeds if it ends the episode within a fixed radius of the goal. In our experiments we use a success radius of 0.2 meters and the agent is spawned anywhere from 1 to 30 meters from the goal. The agent is provided with a one-hot encoding of its previous action. Since the agent is given the distance to the goal, learning the Stop action is trivial, so we disregard it.

\textbf{Scene Exploration (Exploration)}
\label{sec:exploration}
We discretize the world-space into 1 meter cubes and count the number of distinct cubes visited by the agent during a fixed duration. This task differs from Point-to-Point Navigation in that no absolute or relative spatial locations are provided to the agent. This prohibits agents from learning to detect collisions by comparing location values from two timesteps, requiring them to visually detect collisions. The agent still receives a one-hot encoding of its previous action.

\textbf{Run Away from Location (Flee)}
\label{sec:exploration}
The goal of this task is to maximize the geodesic distance from the start location and the agent's final location in episodes of fixed length. As in Exploration, no spatial locations are given to the agent.

\subsection{Decomposing the Learning Problem}

\begin{figure}[t]
  \begin{center}
  \includegraphics[width=\linewidth]{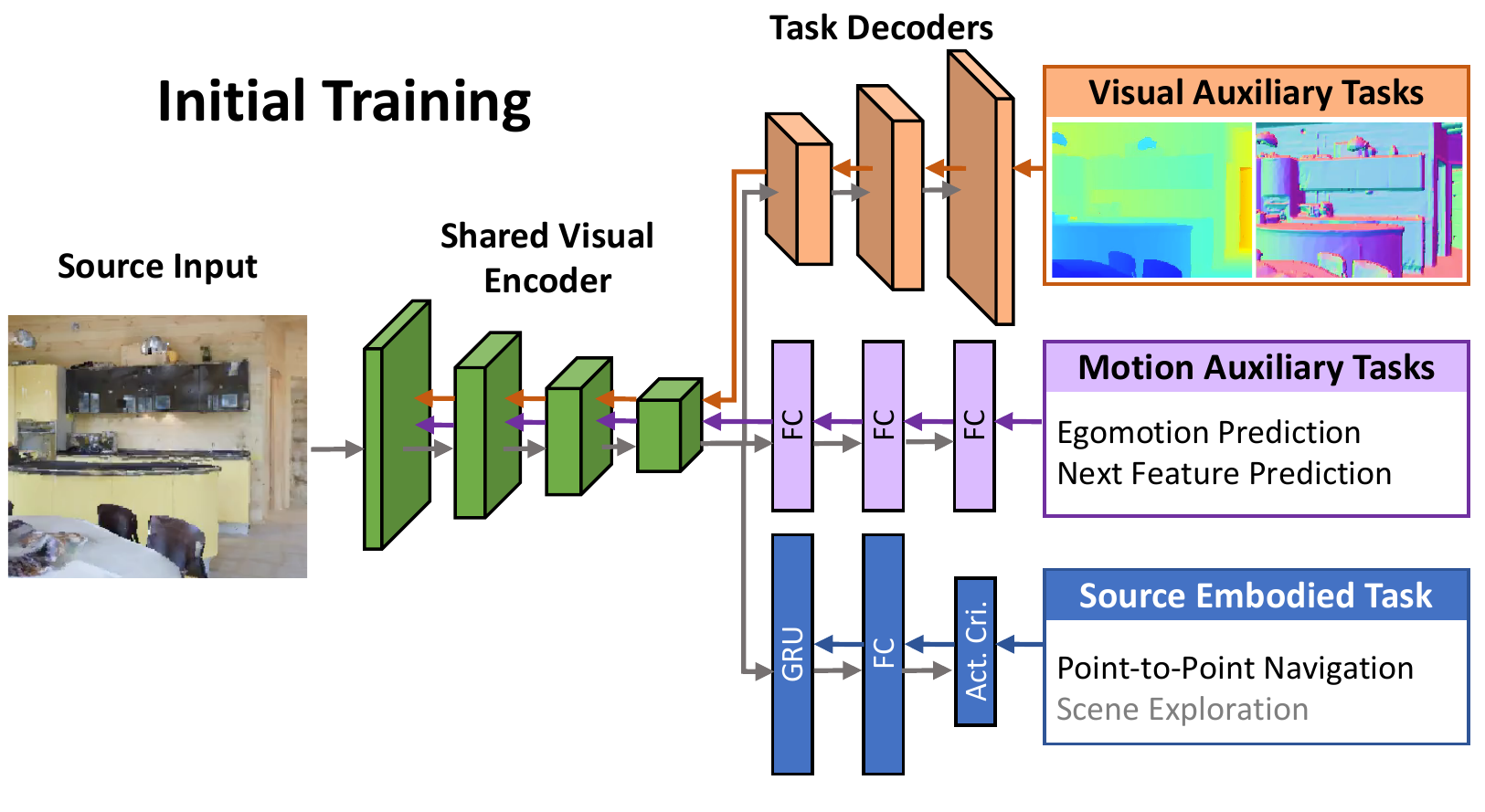}
  \end{center}
  \caption{\textbf{\model initial learning on source data and task.} Given source visual inputs, the visual encoder is trained using auxiliary visual and motion based tasks. Next, the policy decoder is trained on the source embodied tasks with a fixed visual encoder. Gradients from the embodied task (depicted as blue arrows) are stopped before the shared visual encoder to ensure decoupling of the policy and perception.}
  \vspace{-5mm}
  \label{fig:init-train}
\end{figure}
For visual navigation tasks, an agent must \textbf{understand what it sees} and it must use the perceived world to \textbf{decide what to do}. Thus, we decompose visual navigation into the subtasks of (1) encoding the visual information and (2) using the encoded information to navigate. At each time $t$ the agent receives an egocentric image $\image_t$ from the environment and must return a navigation action $\action_t$ in order to accomplish the task. Instead of learning actions directly from pixels, we break the decision-making into two stages. First, a function $\encoder$ processes the image $\image_t$ producing a feature embedding $\visft_t = \encoder(\image_t)$. Next, the features are decoded into an action $\action_t = \decoder(\visft_t)$. Our goal is to learn features $\visft_t$ which extract salient information for completing navigation tasks and which generalize to new environments. Rather than passively expecting the end-to-end training to result in transferable features, we directly optimize  portions of the network with distinct objectives to produce representations which are highly semantically meaningful and transferable. 

\subsection{Visual Encoder}
Visual understanding comes in many forms and is highly dependent on the desired end task. In the case of visual navigation, the agent must convert pixel inputs into an implicit or explicit geometric understanding of the environment's layout. To encapsulate these ideas, we train a bottleneck encoder-decoder network supervised by several auxiliary visual and motion tasks. Each task uses a shared encoder, 
and produces a general purpose feature, $\visft_t$. 
This feature is then used as input to learn a set of task specific decoders. 

\vspace{-3mm}
\paragraph{Auxiliary Visual Tasks:} 
We encourage the shared encoder to extract geometric information from the raw visual input by augmenting the learning objective with the following auxiliary visual tasks: (1) prediction of depth through a depth decoder, $\depth$, (2) prediction of surface normals through a surface normal decoder, $\surfnorm$, and (3) RGB reconstruction through a reconstruction decoder, $\recon$ (sample outputs are shown in the supplementary material).
For an input image $\image_t$ with ground truth depth, $D_t$ and ground truth surface normals $S_t$, the learning objective for each of these auxiliary visual decoders is as follows:
\vspace{-1mm}
\begin{align}
    \mathcal{L}_{D} &= \sum_{pixels} \norm{\depth(\visft_t) - D_t}_{1} \\
    \mathcal{L}_{S} &=  1 - \sum_{pixels} \frac{\surfnorm(\visft_t) \cdot S_t}{\norm{\surfnorm(\visft_t)}_2 * \norm{S_t}_2} \\
    \mathcal{L}_{R} &= \sum_{pixels} \norm{\recon(\visft_t) - \image_t}_{1}
\end{align}     
\vspace{-3mm}

\noindent We use the $\ell_1$ loss for reconstruction and depth to encourage edge sharpness. We use the cosine loss for the surface normals as it is a more natural fit for an angular output. 

\vspace{-3mm}
\paragraph{Auxiliary Motion Tasks:} 
We additionally encourage the visual encoder to extract information which may be generically useful for future embodied tasks by adding the following auxiliary motion tasks: (1) predict the egomotion (discrete action) of the agent with motion decoder $\ego$, and (2) forecast the next features given the current features and a one-hot encoding of the action performed with motion decoder $\predNext$. For a visual encoding $\visft_t$ at time $t$, previous encoding $\visft_{t-1}$, and action $a_t$ that causes the agent to move from $\image_{t-1}$ to $\image_t$, the learning objective for each of these auxiliary motion decoders is as follows:
\begin{align}
    \mathcal{L}_{E} &= -\sum_{a \in A} p(a_t = a) \log(\ego(\visft_{t}, \visft_{t-1})) \\
    \mathcal{L}_{P} &= 1 - \sum_{features} \frac{\predNext(\visft_{t-1}, \action_t) \cdot \visft_t}{\norm{\predNext(\visft_{t-1}, \action_t)}_2 * \norm{\visft_t}}_2
\end{align}

\noindent We use the cross-entropy loss as we use a discrete action space, and we use the cosine loss for next feature prediction as it directly normalizes for scale which stops the network from forcing all the features arbitrarily close to 0. 

All objectives affecting the learning of the visual encoder can be summarized in the joint loss:

 $\mathcal{L} = \lambda_{R} \mathcal{L}_{R} + \lambda_{D} \mathcal{L}_{D} + \lambda_{S} \mathcal{L}_{S} + \lambda_{E} \mathcal{L}_{E} + \lambda_{P} \mathcal{L}_{P}$

\noindent where $\lambda_{R}, \lambda_{D}, \lambda_S, \lambda_E, \lambda_P$ are scalar hyperparameters which control the trade-off between the various tasks in this multi-task learning objective.

Rather than expecting our network to learn to extract geometric information decoupled from the policy decoders, we force the visual representation to contain this information directly. This decreases the likelihood of overfitting to training environments and thus increases the likelihood that our model generalizes to unseen environments.

\vspace{-1mm}
\subsection{Policy Decoder}
\vspace{-2mm}
Our policy decoder takes as input the visual features $\visft_t$ and learns to predict a desired action, $\action_{t+1}$, supervised by a reward signal provided by the desired task. To avoid purely reactive policies, we employ a GRU~\cite{gru} to add temporal context. The output of the policy layers predicts a probability distribution over the discretized action space and a value estimate for the current state. The probability distribution is sampled to determine which action to perform next.

When training the policy decoder, we fix our visual encoder and optimize only the policy decoder weights for the chosen task i.e. \textbf{gradients do not propagate from the source task to the visual layers} (see Figure~\ref{fig:init-train} for an illustration of the gradient flow from the embodied task loss). This prevents policy information from leaking into the visual representation, ensuring the visual encoder generalizes well for many tasks. For the task of Point-to-Point Navigation we use two training strategies: \textbf{BC} and \textbf{BC, PPO}.

\noindent \textbf{BC:} We train the agent using behavioral cloning (BC) where the ground truth represents the action which would maximally decrease the geodesic distance between the current position and the goal.
This is trained in a ``student-forcing'' regime i.e. the agent executes actions based on its policy, but evaluates the actions using the ground truth. 

\noindent \textbf{BC, PPO:} We initialize the agent with the weights from the BC setting and update only the policy layers using the PPO algorithm~\cite{ppo} with a shaped reward based on the geodesic distance to the goal, $Geo(P, G)$:
\begin{align}
 r^{pointnav}_t = Geo(P_{t-1}, G) - Geo(P_t, G) + \lambda_{T}
\end{align}
\noindent where $P_t$ is the agent's location at time $t$, $G$ is the goal location, and $\lambda_{T}$ is a small constant time penalty.
\subsection{Selective Transfer to New Domains and Tasks}

\subsubsection{Adapting to new Visual Domains}
\label{sec:method_sim2sim}
\begin{figure}[t]
  \begin{center}
  \includegraphics[width=\linewidth]{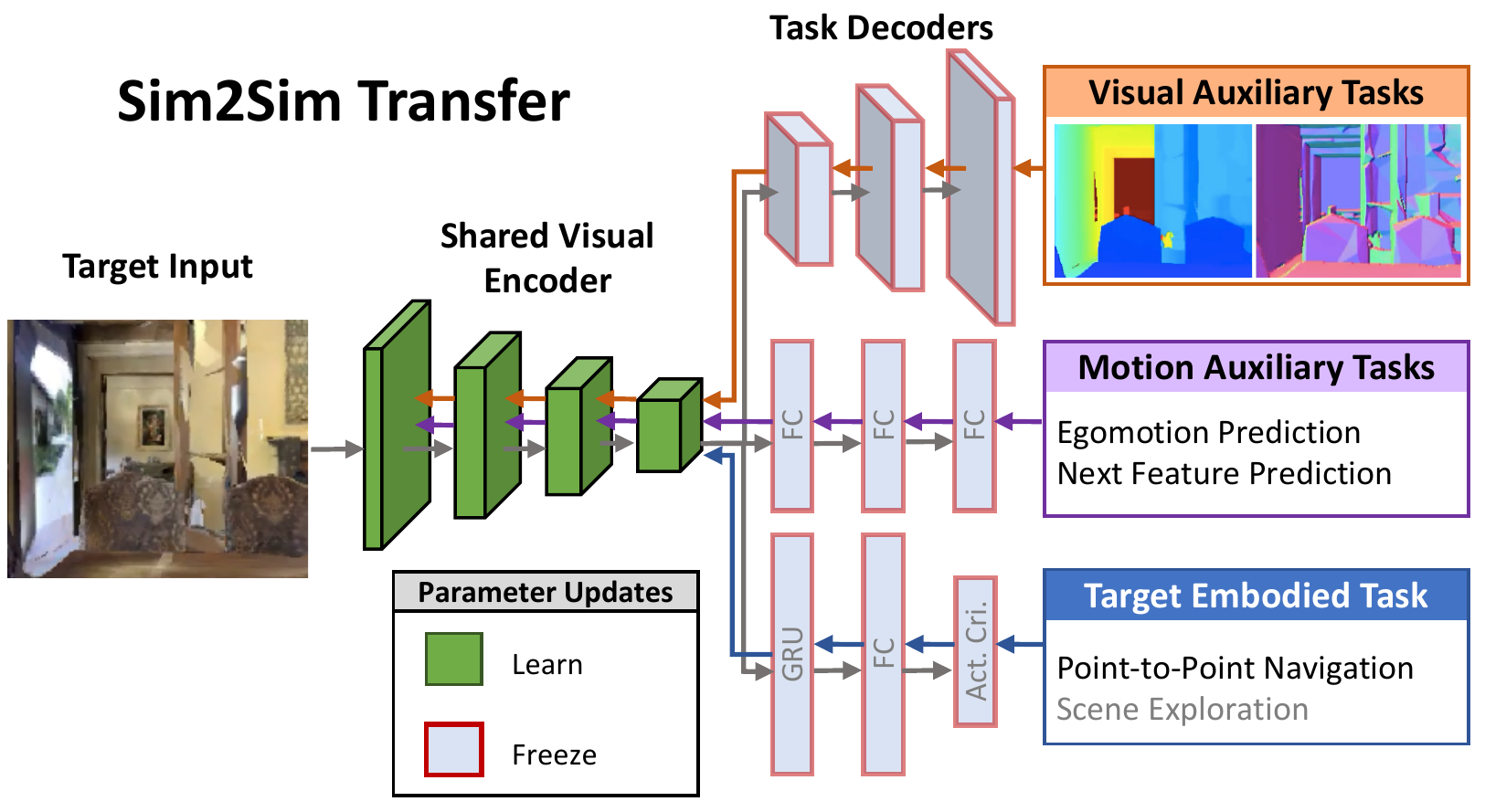}
  \end{center}
  \caption{\textbf{\model{} visual domain transfer.} When the target visual inputs differ from the source visual inputs while the desired embodied task remains fixed, our model updates the shared visual encoder using only auxiliary visual and motion based learning tasks. All decoder weights are frozen (to prevent overfitting), but gradients propagate through all decoder layers to the encoder.}
  \label{fig:domaintransfer}
\end{figure}
By decomposing the learning task into a perceptual encoder and a policy decoder, each supervised by their own objectives, our model is able to learn more transferable visual features than end-to-end methods. Furthermore, our model can quickly adapt its perceptual understanding with auxiliary visual and motion based training in the target environment without needing to modify the policy. Figure~\ref{fig:domaintransfer} illustrates the visual encoder adaptation learning procedure. 
Given a small sample of data and tasks in the target domain, we backpropagate gradients through the policy decoder\footnote{Without propagating gradients through the policy decoder, the encoder feature representation shifts and no longer matches the policy decoder.} and the auxiliary task layers \textbf{but freeze the weights for all but the shared visual encoder}. By doing so, our model can quickly adapt its perception without overfitting the policy to the small sample.

\vspace{-3mm}
\subsubsection{Adapting to new Tasks}
\begin{figure}[t]
  \begin{center}
  \includegraphics[width=\linewidth]{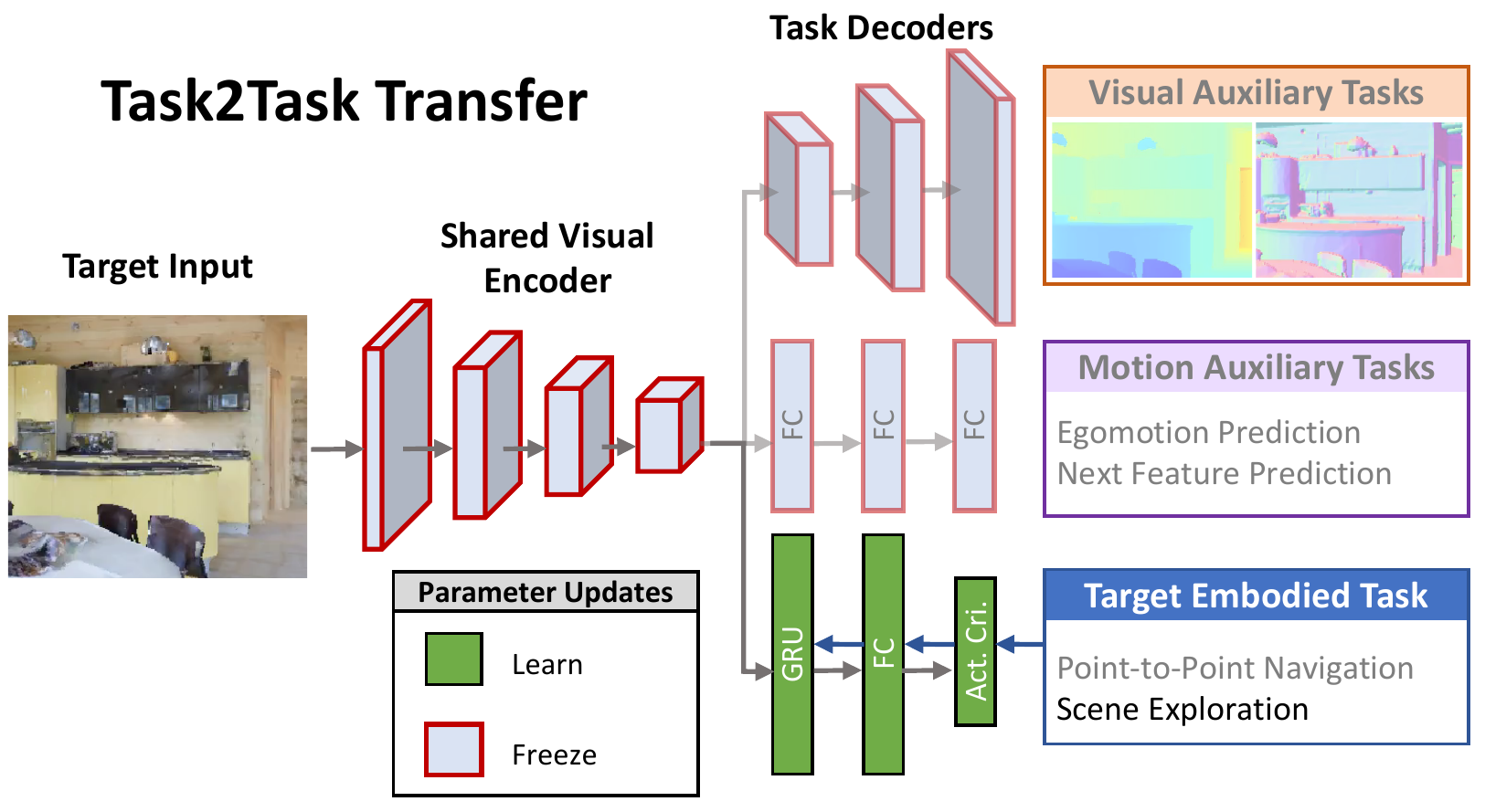}
  \end{center}
  \caption{\textbf{\model{} task transfer.} When learning a new embodied task for the same visual inputs as in the source initial learning, our model fixes the shared visual encoder and updates the policy decoder using the new target embodied loss.}
  \label{fig:tasktransfer}
\end{figure}

When transferring to a new embodied task operating in the same visual space, our model only needs to update the policy decoder parameters (see Figure~\ref{fig:tasktransfer}).
While reusing lower-level features for new tasks by replacing and retraining the final layers is a common technique in deep learning~\cite{2014cvpr_rcnn,2015fcn,balanced_vqa_v2} our model naturally decouples perception and reasoning offering a clear solution as to which layers to freeze or finetune. Rich perceptual features often transfer to tasks which require different reasoning assuming that the representation encodes the necessary information for the new task. By using auxiliary tasks to inform the updates to the visual encoder, we aim to encourage learning of intermediate features that capture semantically meaningful information which should better transfer to new tasks than arbitrary latent features. For example, the latent features from a purely end-to-end learned model may represent a variety of different (sometimes spurious) correlations, while our features must contain enough information to reconstruct depth and surface normals etc., so the network should be able to, for instance, avoid obstacles using the exact same features.
While this implies that the selection of an appropriate auxiliary task affects the success of our method, if necessary our network can still be trained end-to-end using the pretrained weights as initialization.

In the specific cases of transferring from Point-Nav to Exploration or Flee we initialize the model with the weights from the \textbf{BC, PPO} setting and update the policy decoder layers using PPO with the new reward functions:
\begin{align}
    r^{explore}_t &= \norm{Visited_t} - \norm{Visited_{t - 1}} + \lambda_{T} \\
    r^{flee}_t &= Geo(P_{t}, P_{t_0}) - Geo(P_{t-1}, P_{t_0}) + \lambda_{T}
\end{align}

where $\norm{Visited_t}$ represents how many unique spatial locations the agent has visited at time $t$.

\section{Experiments}
To evaluate visual navigation tasks we use the Habitat scene renderer~\cite{habitat19arxiv} on scenes from the near-photo-realistic 3D room datasets Matterport 3D (referred to as MP3D)~\cite{mp3d} and Gibson~\cite{gibson} as well as a third 3D navigation dataset (referred to as \thirddataset{}).

\begin{figure*}[!b]
\begin{center}
\includegraphics[width=0.8\linewidth]{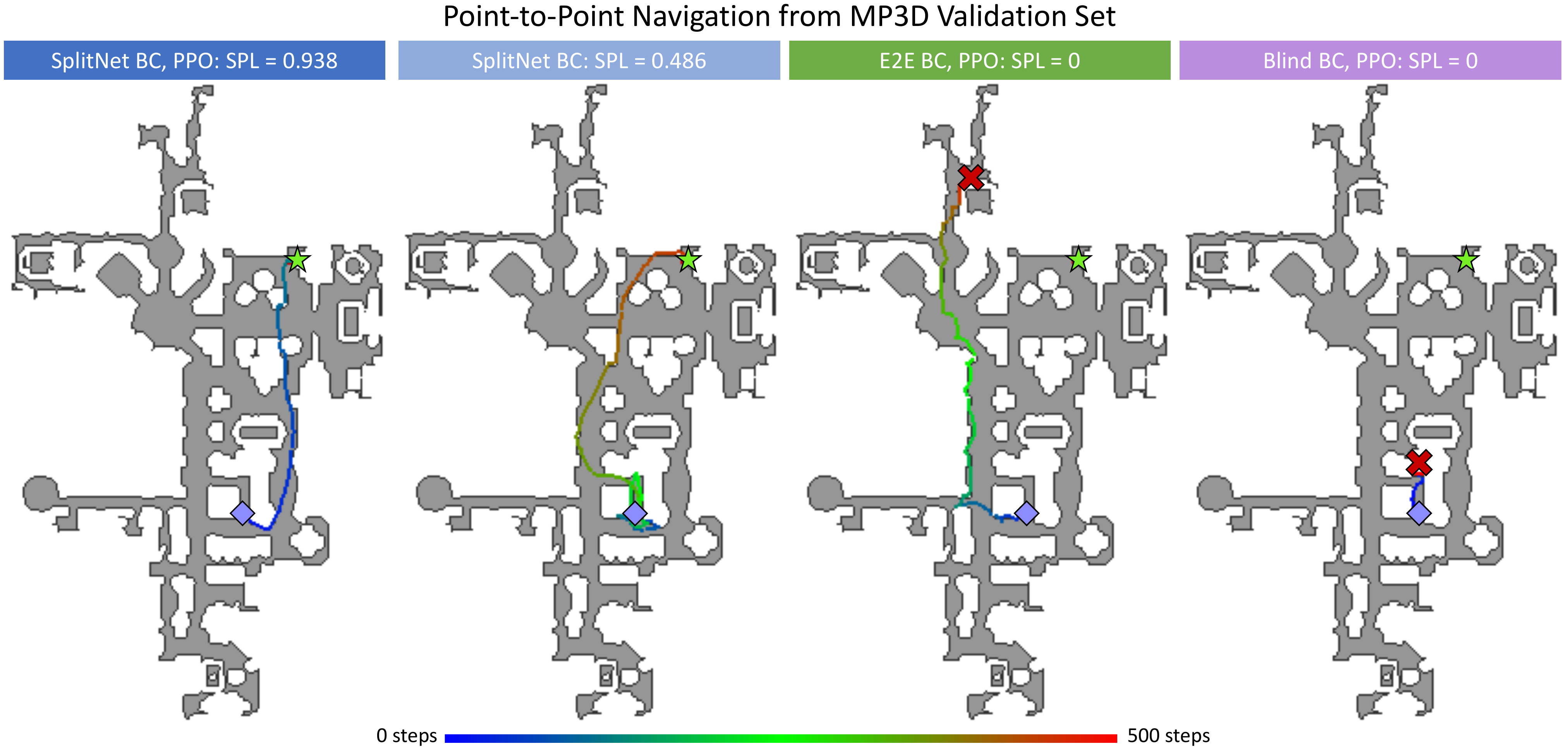}
\caption{\textbf{Qualitative comparison of Point-Nav policies on MP3D validation.} An exemplar validation episode (fixed start and end location) and the predicted trajectories from baselines and \model. 
}
\label{fig:top_down}
\end{center}
\end{figure*}

\subsection{Baselines}
We compare our results against traditional end-to-end (E2E) training algorithm results for all experiments. These can be trained via the PPO algorithm, via behavioral cloning (BC), or pretrained with BC and finetuned with PPO. One common technique across deep learning is to pretrain models on ImageNet~\cite{imagenet}, and finetune the entire network on the desired task, which we also include as a baseline. We do not freeze any weights when training E2E methods. We additionally include blind (but learned) agents for each task and random-action agents to benchmark task difficulty. For Point-Nav, we also include a Blind Goal Follower which aligns itself in the direction of goal vector and moves forward, realigning after it collides with obstacles.

\subsection{Generalization to Unseen Environments}
The ability for an algorithm to generalize to unseen environments represents its effectiveness in real-world scenarios. To begin analyzing our model, we experiment with the standard protocol of training and evaluating on data from the same simulator, partitioning the scenes into train and test. We compare performance for the Point-Nav task on three simulators (\thirddataset{}, MP3D~\cite{mp3d}, Gibson~\cite{gibson}) evaluating in never-before-seen scenes. We use the SPL metric proposed in~\cite{anderson2018evaluation} which can be stated as 
\vspace{-2mm}
\begin{equation}
    SPL = \frac{1}{N}\sum_{i=1}^N{S_i \frac{\ell_i}{\text{max}(p_i, \ell_i)}}
    \vspace{0mm}
\end{equation}
\noindent where $S_i$ is a success indicator for episode $i$, $p_i$ is the path length, and $\ell_i$ is the shortest path length. This combines the accuracy (success) of a navigation method with its efficiency (path length) where 1.0 would be an oracle agent.

\begin{table}
\begin{center}
\resizebox{\columnwidth}{!}{
\begin{tabular}{@{}l cc cc cc@{}} 
\toprule
                                   & \multicolumn{2}{c}{\textbf{\thirddataset{}}}                   & \multicolumn{2}{c}{\textbf{MP3D}~\cite{mp3d}}       & \multicolumn{2}{c}{\textbf{Gibson}~\cite{gibson}}       \\
                                   \cmidrule(lr){2-3}  \cmidrule(lr){4-5}  \cmidrule(lr){6-7}
                                   & SPL                  & Success               & SPL            & Success        & SPL            & Success         \\ 
\midrule
Random                             & 0.012 & 0.027 & 0.011          & 0.016          & 0.046          & 0.028           \\
Blind Goal Follower                & 0.199 & 0.203 & 0.155          & 0.158          & 0.325          & 0.319           \\
Blind BC                   & 0.159                & 0.323                 & 0.232          & 0.382          & 0.351          & 0.603           \\
Blind BC, PPO              & 0.291        & 0.377         & 0.317          & 0.471          & 0.427               & 0.643    \\
Blind PPO                          & 0.258        & 0.371         & 0.313          & 0.463          & 0.538          & 0.822           \\
E2E PPO                     & 0.324        & 0.529         & 0.322          & 0.477          & 0.634          & 0.831           \\
E2E BC              & 0.343                & 0.548                 & 0.459          & 0.737          & 0.509          & 0.824           \\
E2E BC, PPO & 0.393                & 0.593                 & 0.521          & 0.733          & 0.606          & {0.869}  \\
ImageNet Pretrain, E2E BC  & 0.280 &  0.499 &  0.315 &  0.552 &  0.548 &  0.843 \\
ImageNet Pretrain, E2E BC, PPO & 0.338 & 0.440 & 0.450 & 0.539 & 0.642 & 0.737 \\
\model{} BC                    & 0.421                & 0.687                 & 0.517          & 0.808          & 0.584          & 0.865           \\
\model{} BC, PPO                  & \textbf{0.560}       & {0.703}        & \textbf{0.716} & {0.844} & \textbf{0.701} & 0.855           \\
\bottomrule
\end{tabular}
}
\end{center}
\caption{\textbf{Performance on Unseen Environments.} Blind methods are not provided with visual input but still receive an updated goal vector. ``BC, PPO'' methods are first trained with a softmax loss to take the best next action and are finetuned with the PPO algorithm.}
\label{table:generalization}
\vspace{-5mm}
\end{table}

Effective policies generalize by understanding the geometry of the scenes rather than trying to localize into a known map based on the visual inputs. \model{} outperforms all other methods by a wide margin on all three environments (shown in Table~\ref{table:generalization}). Surprisingly, pretraining on ImageNet does not offer better generalization, likely because the features required for ImageNet are sufficiently different from those needed to navigate effectively (note, the convolutional weights trained on ImageNet \textit{are not frozen} during BC and PPO training). This is true even compared to E2E without pretraining on ImageNet.

\begin{table*}[t]
\begin{center}
\resizebox{\textwidth}{!}{
\begin{tabular}{@{}l cccc  cccc@{}} 
\toprule
                    & \multicolumn{2}{c}{\textbf{Number Train Scenes}}        & \multicolumn{2}{c}{\textbf{Test Data}}  & \multicolumn{2}{c}{\textbf{Number Train Scenes}}        & \multicolumn{2}{c}{\textbf{Test Data}}  \\

         & \thirddataset{}                & MP3D (Train)         & \multicolumn{2}{c}{MP3D (Val)}              & MP3D                 & Gibson (Train)       & \multicolumn{2}{c}{Gibson (Val)}             \\
\cmidrule(lr){2-3} \cmidrule(lr){6-7} \cmidrule(lr){4-5} \cmidrule(lr){8-9}
                    & (\textit{Source}) & (\textit{Target}) & SPL                  & Success               & (\textit{Source}) & (\textit{Target}) & SPL                  & Success               \\ 
\midrule
Source E2E BC, PPO     & 990                  & 0                    & 0.257                & 0.412                 & 61                   & 0                    & 0.609                & {0.866}        \\
Source \model{} BC, PPO               & 990                  & 0                    & \textbf{0.376}       & {0.539}        & 61                   & 0                    & \textbf{0.651}       & 0.764                 \\ 
\midrule
Target E2E BC      & 0                    & 1                    & 0.211                & 0.321                 & 0                    & 1                    & 0.396                & 0.589                 \\
Target E2E Finetune      & 990                  & 1                    & Failure             & Failure              & 61                   & 1                    & Failure             & Failure              \\
Target \model{} Transfer     & 990                  & 1                    & \textbf{0.447}       & {0.596}        & 61                   & 1                    & \textbf{0.686}       & {0.822}        \\ 
\midrule
Target E2E BC      & 0                    & 10                   & 0.259                & 0.463                 & 0                    & 10                   & 0.501                & 0.782                 \\
Target E2E Finetune     & 990                  & 10                   & 0.401                & 0.612                 & 61                   & 10                   & 0.667                & {0.870}        \\
Target \model{} Transfer      & 990                  & 10                   & \textbf{0.531}       & {0.681}        & 61                   & 10                   & \textbf{0.727}       & 0.854                 \\ 
\midrule
Target E2E BC, PPO     & 0                    & All (61)             & 0.521                & 0.733                 & 0                    & All (72)             & 0.606                & {0.869}        \\
Target \model{} BC, PPO                & 0                    & All (61)             & \textbf{0.716}       & {0.844}        & 0                    & All (72)             & \textbf{0.701}       & 0.855                 \\
\bottomrule
\end{tabular}
}
\end{center}
\caption{\textbf{Performance transferring across simulation environments (Sim2Sim)}. Our method, \model, significantly outperforms the end-to-end (E2E) baseline at the task of transferring across simulated environments. For reference, we also report the performance of a source only trained model (top two rows) or a target only trained model (bottom two rows). ``Failure'' indicates that performance on the target data decreases after finetuning.}
\label{table:transfer}
\end{table*}

As qualitative intuition about the performance of the various methods, we depict the policies for a subset of methods on an example MP3D episode from in Figure~\ref{fig:top_down}. For a fixed start (blue diamond) and goal (green star) location, we show the output trajectory from each method where the trajectory color (ranging from blue to red) denotes the number of steps so far. If a policy failed to reach the goal, the final destination is denoted with a red ``x''. From this visualization we can see that \model using BC and PPO successfully completes the task and does so with the shortest overall path. At the beginning of the episode \model{} BC is stuck behind the wall, but eventually is able to navigate away from the wall and reach the target. 

\begin{figure}
\begin{center}
\includegraphics[width=1.0\columnwidth]{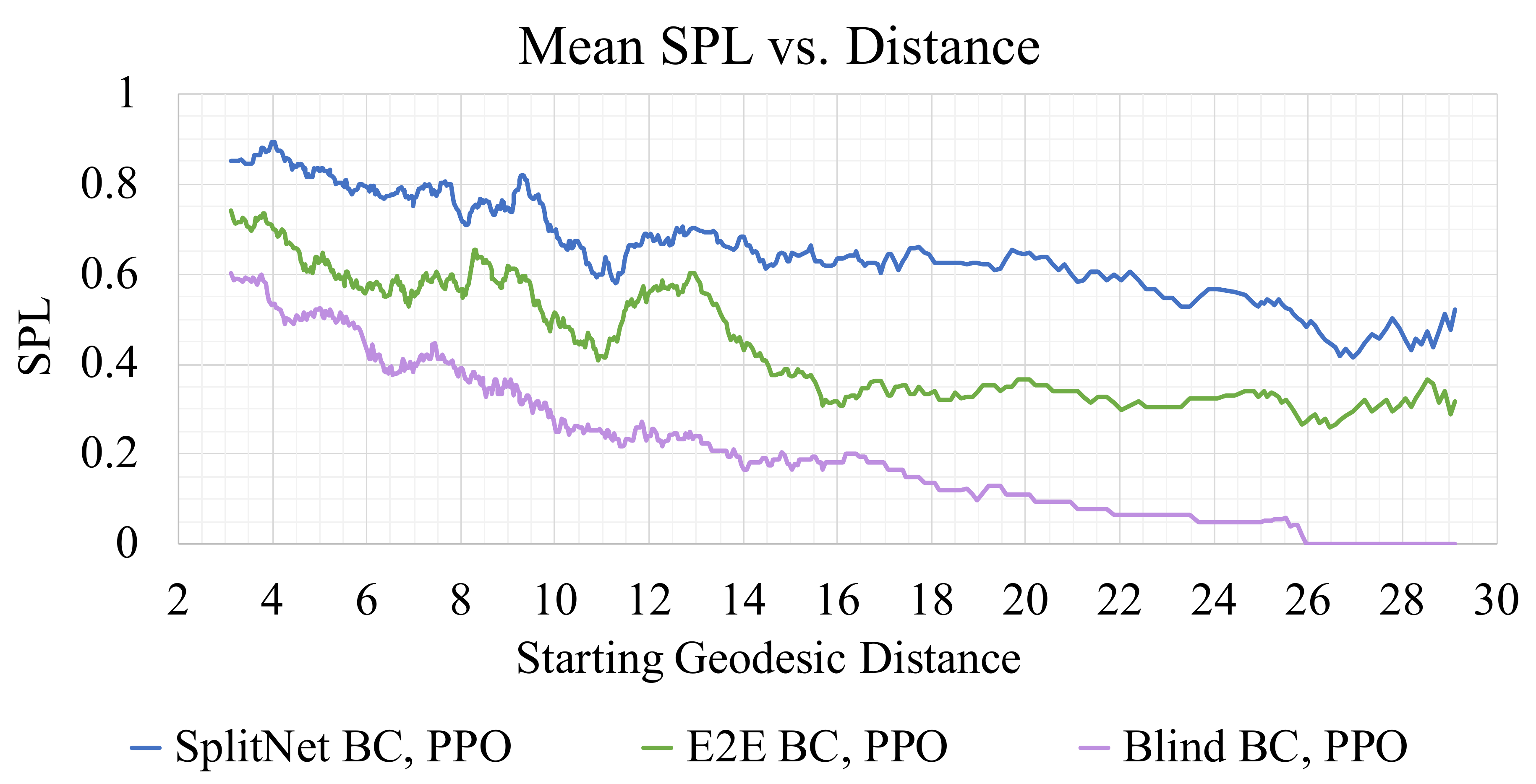}
\caption{\textbf{MP3D Point-Nav Performance vs episode difficulty.} We compare our method, \model, to end-to-end (E2E) and blind learned baselines and report SPL performance as a function of starting geodesic distance from the goal. \model outperforms on all starting distances, especially on the more difficult episodes.
}
\label{fig:distance_plot}
\end{center}
\vspace{-8mm}
\end{figure}

We further analyze the performance of \model compared to baselines as a function of the geodesic distance between the starting and goal locations in Figure~\ref{fig:distance_plot}. This distance is highly correlated with the difficulty of an episode. 
Unsurprisingly, all methods degrade as the starting location is moved further from the goal location, but \model retains its advantage over baselines irrespective of the episode difficulty. Additionally, we see the performance gap widen over the more difficult episodes, meaning we handle difficult episodes better than the baselines.

\subsection{Transfer Across Simulators}
We now study the ability for our method to transfer between visual environments for the fixed task of Point-Nav. 
We denote \textit{Source} to be the initial simulator in which we train our model using both BC and PPO and denote this initial model as ``Source \model BC, PPO.''  The baseline source model that uses end-to-end training is denoted as ``Source E2E BC, PPO.''
We then compare our method for transfer to the new simulator \textit{Target}, described in Section~\ref{sec:method_sim2sim} and denoted as ``Target \model Transfer,'' against the end-to-end baseline finetuned on the target, ``Target E2E Finetune.'' For reference, we also present the performance of training an end-to-end model using only the available target data, denoted as ``Target E2E BC.''

Table~\ref{table:transfer} reports our main results for this Sim2Sim transfer problem as a function of the amount of available target scenes during training. We report performance for the two transfer settings of \thirddataset{}$\rightarrow$MP3D and MP3D$\rightarrow$Gibson. 
%
These simulators differ in terms of complexities of differing rendering appearance (as seen in Figure~\ref{fig:teaser}), different environment construction methods (synthetic vs. depth-scan reconstruction), and different environment size. 
Again, \model{} outperforms all baselines across all experiments in terms of  the SPL metric and performs better or comparable to the baseline in terms of success for all transfer setups. 
Even with no extra data, our initially learned network is more generalizable to new environments, especially those which are significantly different in appearance (\thirddataset{}$\rightarrow$MP3D). 
Of note, in both cases, \model{} given 10 scenes from the target dataset matches or outperforms the end-to-end baseline SPL given the entire target dataset.

Note, that our approach to visual environement transfer includes finetuning only the visual encoder in the target environment and leaving the policy decoder fixed. One may wonder whether this is the optimal approach or whether our method would benefit from target updates to the policy decoder as well. To answer this question, in Table~\ref{table:finetune_ablation} we report performance comparing the initial source \model performance to that of finetuning either only the visual encoder (``V'') which is our proposed approach or finetuning both the visual encoder and policy decoder (``V+P''). Interestingly, we found that allowing updates to both the visual encoder and policy decoder in the target environment lead to significant overfitting which resulted in failed generalization to the unseen scenes from the validation sets. This confirms the benefit of our split training approach.
\begin{table}[t]
    \begin{center}
    \resizebox{\linewidth}{!}{
    \begin{tabular}{@{}lcc  cc@{}}
        \toprule
         \multirow{2}{*}{\textbf{\model Model}} & \textbf{Layers} & \textbf{Number} &  \multirow{2}{*}{\textbf{SPL}} &  \multirow{2}{*}{\textbf{Success}} \\
          & \textbf{Finetuned} & \textbf{Target Scenes}\\
         \midrule
         \multicolumn{5}{c}{\textit{Transfer \thirddataset{} $\rightarrow$ MP3D (train): Eval MP3D (val)}}\\
         \midrule
         Source Only & -                & -                    & {0.376}    & {0.539}\\
         Finetune Target & V+P    & 1 &    0.435             & 0.586 \\
         Finetune Target & V        & 1 &   \textbf{0.447}    & {0.596} \\
         \midrule
         Finetune Target & V+P    & 10 & 0.400             & 0.552     \\
         Finetune Target & V        & 10 &  \textbf{0.531}    & {0.681}  \\
         \midrule 
         \multicolumn{5}{c}{\textit{Transfer MP3D $\rightarrow$ Gibson (train): Eval Gibson (val)}}\\
         \midrule
         Source Only & -                & -                &  {0.651}       & 0.764\\
         Finetune Target & V+P        & 1 &      Failure & Failure\\
         Finetune Target & V        & 1 &     \textbf{0.686}       & {0.822}  \\
         \midrule
         Finetune Target & V+P        & 10 &   Failure & Failure\\
         Finetune Target & V        & 10 &    \textbf{0.727}       & {0.854}   \\
         \bottomrule
    \end{tabular}
    }
    \end{center}
    \caption{\textbf{Ablation of \model Sim2Sim transfer strategy}. \model only updates the visual encoder (``V'') and fixes the policy decoder (``P'') when finetuning the source \model model on a target visual environment. In contrast, finetuning both V+P on the target leads to degraded performance.}
    \label{table:finetune_ablation}
    \vspace{-5mm}
\end{table}

\subsection{Transfer Across Tasks}
We test the ability for \model{} to learn a new task by first training the network on Point-Nav and using our approach to transfer the model to the novel tasks of Exploration and Flee. All three tasks require the ability to transform 2D visual inputs into 3D scene geometry, but the decisions about what to do based on the perceived geometry are drastically different. Since \model{} decouples the policy from the perception, it learns features which readily transfer to a novel, but related task. 

\begin{figure}[t]
\begin{center}
\includegraphics[width=1.0\columnwidth]{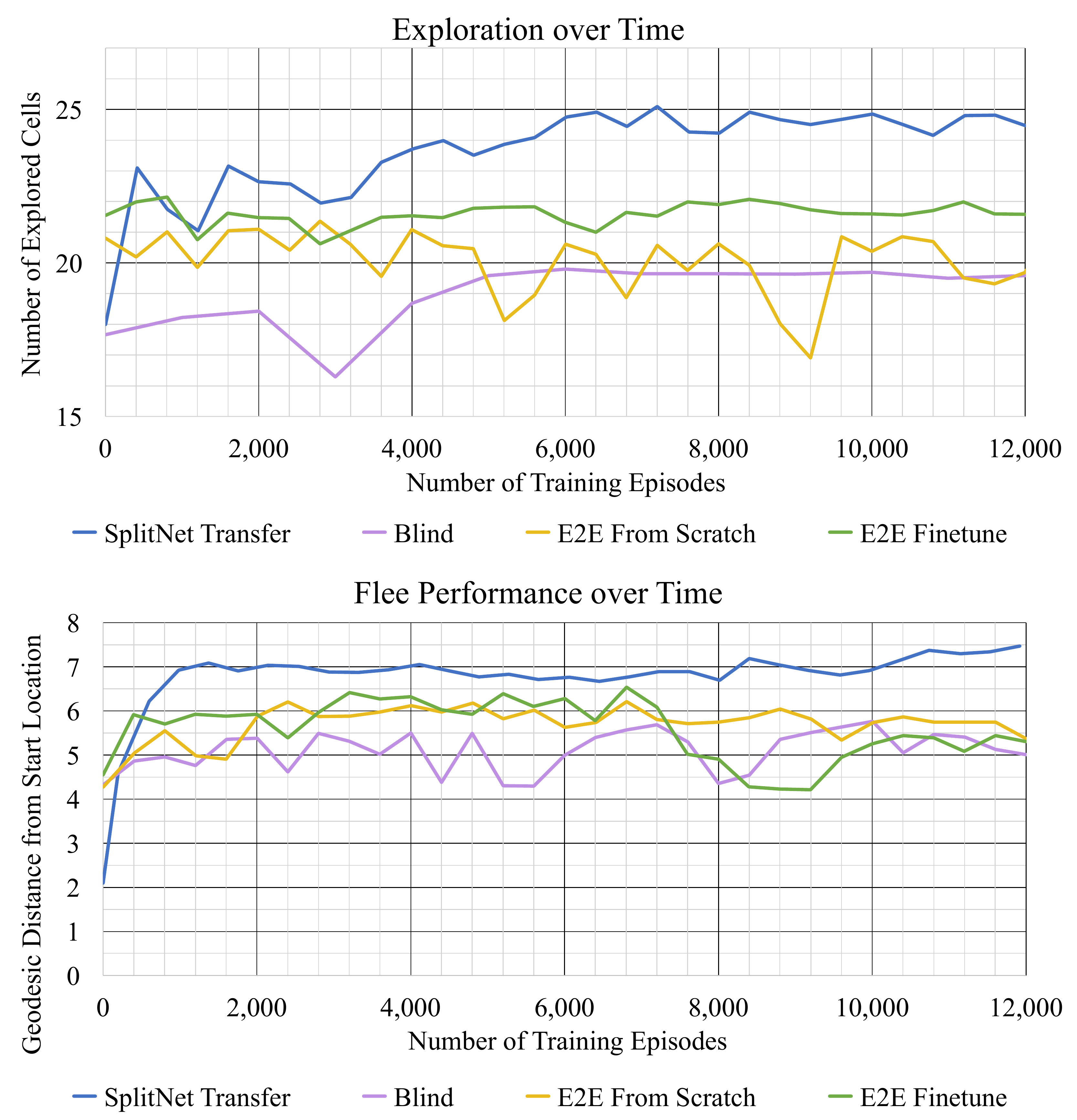}
\caption{\textbf{\thirddataset{} Task2Task performance as a function of target training episodes.} \model{} Transfer and E2E Transfer are first trained on \thirddataset{} Point-Nav, but \model{} only updates the policy layers whereas E2E updates the entire network. E2E from scratch is randomly initialized a episode 0. The Blind method only receives its previous action as input and is randomly initialized. Oracle agents perform at 33.5 and 19.5 respectively.
}
\label{fig:task2task}
\end{center}
\vspace{-8mm}
\end{figure}

Figure~\ref{fig:task2task} shows that \model{} immediately begins to learn effective new policies, outperforming the other methods almost right away. In Exploration, our method is able to reuse its understanding of depth to quickly learn to approach walls, then turn at the last second and head off in a new direction. For the Flee task, our method identifies long empty hallways and navigates down those away from the start location. None of the other methods learn robust obstacle-avoidance behavior or geometric scene understanding.
Instead they latch on to simple dataset biases such as ``repeatedly move forward then rotate.'' Oracle agents perform at 33.5 and 19.5 respectively, but are not directly comparable in that they benefit from knowing the environment layout before beginning the episode. 

\subsection{Analysis of Auxiliary Objectives}
Our method was designed as a solution for generalization on a downstream embodied task. 
However, \model{} also learns outputs for the auxiliary visual and motion tasks. While our goal is not to surpass state-of-the-art performance on these auxiliary tasks it is still useful to verify that the visual encodings match our expectations. We therefore include several examples which show the auxiliary outputs in the supplemental material. In our our initial experiments, we found depth and normal estimation to be the most important auxiliary task as they most directly translate to navigation understanding (free space, geometry, etc.). A complete ablation of auxiliary tasks and its consequences is left as future work.

\vspace{-2mm}
\section{Conclusion}
\vspace{-2mm}
We introduce \model{}, a method for decomposing embodied learning tasks to enable fast and accurate transfer to new environments and new tasks. By disentangling the visual encoding of the state from the policy for a task, we learn more robust features which can be frozen or adapted based on the changed domain. Our model matches the performance of end-to-end methods even with six times less data. We believe \model{} may prove to be a useful stepping stone in transferring networks from simulation environments onto robots in the real-world.
\section{Acknowledgements}
We would like to give a special thanks to Oleksandr Maksymets, Manolis Savva, Erik Wijmans, and the rest of the Habitat team for their immense help in using Habitat.

{\small
\bibliographystyle{ieee_fullname}
\bibliography{sim2sim}

\begin{thebibliography}{10}\itemsep=-1pt

\bibitem{anderson2018evaluation}
Peter Anderson, Angel Chang, Devendra~Singh Chaplot, Alexey Dosovitskiy,
  Saurabh Gupta, Vladlen Koltun, Jana Kosecka, Jitendra Malik, Roozbeh
  Mottaghi, Manolis Savva, et~al.
\newblock On evaluation of embodied navigation agents.
\newblock {\em arXiv preprint arXiv:1807.06757}, 2018.

\bibitem{2017cvpr_bousmalis}
Konstantinos Bousmalis, Nathan Silberman, David Dohan, Dumitru Erhan, and Dilip
  Krishnan.
\newblock Unsupervised pixel-level domain adaptation with generative
  adversarial networks.
\newblock In {\em Computer Vision and Pattern Recogntion (CVPR)}, 20117.

\bibitem{mp3d}
Angel Chang, Angela Dai, Thomas Funkhouser, Maciej Halber, Matthias Niessner,
  Manolis Savva, Shuran Song, Andy Zeng, and Yinda Zhang.
\newblock {Matterport3D}: Learning from {RGB-D} data in indoor environments.
\newblock In {\em International Conference on 3D Vision (3DV)}, 2017.

\bibitem{gru}
Kyunghyun Cho, Bart van Merrienboer, Caglar Gulcehre, Dzmitry Bahdanau, Fethi
  Bougares, Holger Schwenk, and Yoshua Bengio.
\newblock Learning phrase representations using rnn encoder--decoder for
  statistical machine translation.
\newblock In {\em Proceedings of the 2014 Conference on Empirical Methods in
  Natural Language Processing (EMNLP)}, pages 1724--1734, 2014.

\bibitem{imagenet}
J. Deng, W. Dong, R. Socher, L.-J. Li, K. Li, and L. Fei-Fei.
\newblock {ImageNet: A Large-Scale Hierarchical Image Database}.
\newblock In {\em Computer Vision and Pattern Recogntion (CVPR)}, 2009.

\bibitem{2015icml_ganin}
Yaroslav Ganin and Victor Lempitsky.
\newblock Unsupervised domain adaptation by backpropagation.
\newblock In David Blei and Francis Bach, editors, {\em Proceedings of the 32nd
  International Conference on Machine Learning (ICML-15)}, pages 1180--1189.
  JMLR Workshop and Conference Proceedings, 2015.

\bibitem{2014cvpr_rcnn}
Ross Girshick, Jeff Donahue, Trevor Darrell, and Jitendra Malik.
\newblock Rich feature hierarchies for accurate object detection and semantic
  segmentation.
\newblock In {\em Computer Vision and Pattern Recogntion (CVPR)}, 2014.

\bibitem{balanced_vqa_v2}
Yash Goyal, Tejas Khot, Douglas Summers{-}Stay, Dhruv Batra, and Devi Parikh.
\newblock Making the {V} in {VQA} matter: Elevating the role of image
  understanding in {V}isual {Q}uestion {A}nswering.
\newblock In {\em Computer Vision and Pattern Recogntion (CVPR)}, 2017.

\bibitem{gupta2017cognitive}
Saurabh Gupta, James Davidson, Sergey Levine, Rahul Sukthankar, and Jitendra
  Malik.
\newblock Cognitive mapping and planning for visual navigation.
\newblock In {\em Computer Vision and Pattern Recogntion (CVPR)}, pages
  2616--2625, 2017.

\bibitem{2018icml_hoffman}
Judy Hoffman, Eric Tzeng, Taesung Park, Jun-Yan Zhu, Phillip Isola, Kate
  Saenko, Alexei~A. Efros, and Trevor Darrell.
\newblock Cycada: Cycle-consistent adversarial domain adaptation.
\newblock In {\em International Conference in Machine Learning (ICML)}, 2018.

\bibitem{fcninwild}
Judy Hoffman, Dequan Wang, Fisher Yu, and Trevor Darrell.
\newblock Fcns in the wild: Pixel-level adversarial and constraint-based
  adaptation.
\newblock {\em CoRR}, abs/1612.02649, 2016.

\bibitem{gam}
Haoshuo Huang1, Qixing Huang2, and Philipp Kr¨ahenb¨uhl.
\newblock Domain transfer through deep activation matching.
\newblock In {\em European Conference on Computer Vision (ECCV)}, 2018.

\bibitem{2015fcn}
{J. Long*}, {E. Shelhamer∗}, and Trevor Darrel.
\newblock Fully convolutional networks for semantic segmentation.
\newblock In {\em Computer Vision and Pattern Recogntion (CVPR)}, 2015.

\bibitem{jaderberg2016reinforcement}
Max Jaderberg, Volodymyr Mnih, Wojciech~Marian Czarnecki, Tom Schaul, Joel~Z
  Leibo, David Silver, and Koray Kavukcuoglu.
\newblock Reinforcement learning with unsupervised auxiliary tasks.
\newblock In {\em International Conference in Learning Representations (ICLR)},
  2017.

\bibitem{ai2thor}
Eric Kolve, Roozbeh Mottaghi, Winson Han, Eli VanderBilt, Luca Weihs, Alvaro
  Herrasti, Daniel Gordon, Yuke Zhu, Abhinav Gupta, and Ali Farhadi.
\newblock {AI2-THOR: An Interactive 3D Environment for Visual AI}.
\newblock {\em arXiv}, 2017.

\bibitem{2016jmlr_levine}
Sergey Levine, Chelsea Finn, and Trevor~Darrell amd Pieter~Abbeel.
\newblock End-to-end training of deep visuomotor policies.
\newblock {\em Journal of Machine Learning Research (JMLR)}, 2016.

\bibitem{levine2018learning}
Sergey Levine, Peter Pastor, Alex Krizhevsky, Julian Ibarz, and Deirdre
  Quillen.
\newblock Learning hand-eye coordination for robotic grasping with deep
  learning and large-scale data collection.
\newblock {\em The International Journal of Robotics Research},
  37(4-5):421--436, 2018.

\bibitem{cogan}
M.-Y. Liu and O. Tuzel.
\newblock Coupled generative adversarial networks.
\newblock In {\em Neural Information Processing Symposium (NeurIPS)}, 2016.

\bibitem{2015icml_long}
Mingsheng Long, Yue Cao, Jianmin Wang, and Michael~I. Jordan.
\newblock Learning transferable features with deep adaptation networks.
\newblock In {\em International Conference in Machine Learning (ICML)}, 2015.

\bibitem{2017iclr_mirowski}
Piotr Mirowski, Razvan Pascanu, Fabio Viola, Hubert Soyer, Andrew~J. Ballard,
  Andrea Banino, Misha Denil, Ross Goroshin, Laurent Sifre, Koray Kavukcuoglu,
  Dharshan Kumaran, and Raia Hadsell.
\newblock Learning to navigate in complex environments.
\newblock In {\em International Conference in Learning Representations (ICLR)},
  2017.

\bibitem{mnih2016asynchronous}
Volodymyr Mnih, Adria~Puigdomenech Badia, Mehdi Mirza, Alex Graves, Timothy
  Lillicrap, Tim Harley, David Silver, and Koray Kavukcuoglu.
\newblock Asynchronous methods for deep reinforcement learning.
\newblock In {\em International conference on machine learning}, pages
  1928--1937, 2016.

\bibitem{2018eccv_drivingsim2real}
Matthias M{\"u}ller, Alexey Dosovitskiy, Bernard Ghanem, and Vladlen Koltun.
\newblock Driving policy transfer via modularity and abstraction.
\newblock In {\em European Conference on Computer Vision (ECCV)}, 2018.

\bibitem{curiosity}
Deepak Pathak, Pulkit Agrawal, Alexei~A Efros, and Trevor Darrell.
\newblock Curiosity-driven exploration by self-supervised prediction.
\newblock In {\em Proceedings of the IEEE Conference on Computer Vision and
  Pattern Recognition Workshops}, pages 16--17, 2017.

\bibitem{peng2018sim}
Xue~Bin Peng, Marcin Andrychowicz, Wojciech Zaremba, and Pieter Abbeel.
\newblock Sim-to-real transfer of robotic control with dynamics randomization.
\newblock In {\em 2018 IEEE International Conference on Robotics and Automation
  (ICRA)}, pages 1--8. IEEE, 2018.

\bibitem{dcgan}
Alec Radford, Luke Metz, and Soumith Chintala.
\newblock Unsupervised representation learning with deep convolutional
  generative adversarial networks.
\newblock In {\em International Conference in Learning Representations (ICLR)},
  2016.

\bibitem{rusu2017sim}
Andrei~A Rusu, Matej Ve{\v{c}}er{\'\i}k, Thomas Roth{\"o}rl, Nicolas Heess,
  Razvan Pascanu, and Raia Hadsell.
\newblock Sim-to-real robot learning from pixels with progressive nets.
\newblock In {\em Conference on Robot Learning (CoRL)}, pages 262--270, 2017.

\bibitem{divis}
Fereshteh Sadeghi.
\newblock {DIViS}: Domain invariant visual servoing for collision-free goal
  reaching.
\newblock In {\em Robotics: Science and Systems(RSS)}, 2019.

\bibitem{sadeghi2017cadrl}
Fereshteh Sadeghi and Sergey Levine.
\newblock {CAD2RL}: Real single-image flight without a single real image.
\newblock In {\em Robotics: Science and Systems(RSS)}, 2017.

\bibitem{sim2real_viewpoint}
Fereshteh Sadeghi, Alexander Toshev, Eric Jang, and Sergey Levine.
\newblock Sim2real viewpoint invariant visual servoing by recurrent control.
\newblock In {\em Proceedings of the IEEE Conference on Computer Vision and
  Pattern Recognition}, pages 4691--4699, 2018.

\bibitem{habitat19arxiv}
Manolis Savva, Abhishek Kadian, Oleksandr Maksymets, Yili Zhao, Erik Wijmans,
  Bhavana Jain, Julian Straub, Jia Liu, Vladlen Koltun, Jitendra Malik, Devi
  Parikh, and Dhruv Batra.
\newblock Habitat: A platform for embodied ai research.
\newblock {\em arXiv preprint arXiv:1904.01201}, 2019.

\bibitem{perceptualactor}
Alexander Sax, Bradley Emi, Amir~Roshan Zamir, Leonidas~J. Guibas, Silvio
  Savarese, and Jitendra Malik.
\newblock Mid-level visual representations improve generalization and sample
  efficiency for learning active tasks.
\newblock {\em CoRR}, abs/1812.11971, 2018.

\bibitem{ppo}
John Schulman, Filip Wolski, Prafulla Dhariwal, Alec Radford, and Oleg Klimov.
\newblock Proximal policy optimization algorithms.
\newblock {\em arXiv preprint arXiv:1707.06347}, 2017.

\bibitem{2016eccv_sun}
Baochen Sun and Kate Saenko.
\newblock Deep coral: Correlation alignment for deep domain adaptation.
\newblock In {\em European Conference on Computer Vision (ECCV)}, 2016.

\bibitem{2017iclr_taigman}
Y. Taigman, A. Polyak, and L. Wolf.
\newblock Unsupervised cross-domain image generation.
\newblock In {\em International Conference in Learning Representations (ICLR)},
  2017.

\bibitem{tobin2017domain}
Josh Tobin, Rachel Fong, Alex Ray, Jonas Schneider, Wojciech Zaremba, and
  Pieter Abbeel.
\newblock Domain randomization for transferring deep neural networks from
  simulation to the real world.
\newblock In {\em 2017 IEEE/RSJ International Conference on Intelligent Robots
  and Systems (IROS)}, pages 23--30. IEEE, 2017.

\bibitem{2017cvpr_tzeng}
Eric Tzeng, Judy Hoffman, Trevor Darrell, and Kate Saenko.
\newblock Adversarial discriminative domain adaptation.
\newblock In {\em Computer Vision and Pattern Recogntion (CVPR)}, 2017.

\bibitem{gibson}
Fei Xia, Amir~R. Zamir, Zhiyang He, Alexander Sax, Jitendra Malik, and Silvio
  Savarese.
\newblock Gibson env: Real-world perception for embodied agents.
\newblock In {\em Computer Vision and Pattern Recogntion (CVPR)}, 2018.

\bibitem{chalet}
Claudia Yan, Dipendra Misra, Andrew Bennnett, Aaron Walsman, Yonatan Bisk, and
  Yoav Artzi.
\newblock {CHALET}: Cornell house agent learning environment.
\newblock {\em arXiv:1801.07357}, 2018.

\bibitem{taskonomy2018}
Amir~R. Zamir, Alexander Sax, William~B. Shen, Leonidas~J. Guibas, Jitendra
  Malik, and Silvio Savarese.
\newblock Taskonomy: Disentangling task transfer learning.
\newblock In {\em Computer Vision and Pattern Recogntion (CVPR)}. IEEE, 2018.

\bibitem{2017iros_zhang}
Jingwei Zhang, Jost~Tobias Springenberg, Joschka Boedecker, and Wolfram
  Burgard.
\newblock Deep reinforcement learning with successor features for navigation
  across similar environments.
\newblock In {\em International Conference on Intelligent Robotics and
  Systems}, 2017.

\bibitem{curriculumDA}
Yang Zhang, Philip David, and Boqing Gong.
\newblock Curriculum domain adaptation for semantic segmentation of urban
  scenes.
\newblock In {\em International Conference on Computer Vision (ICCV)}, 2017.

\bibitem{cyclegan}
Jun-Yan Zhu, Taesung Park, Phillip Isola, and Alexei~A Efros.
\newblock Unpaired image-to-image translation using cycle-consistent
  adversarial networks.
\newblock In {\em International Conference on Computer Vision (ICCV)}, 2017.

\bibitem{zhu2017visual}
Yuke Zhu, Daniel Gordon, Eric Kolve, Dieter Fox, Li Fei-Fei, Abhinav Gupta,
  Roozbeh Mottaghi, and Ali Farhadi.
\newblock Visual semantic planning using deep successor representations.
\newblock In {\em International Conference on Computer Vision (ICCV)}, pages
  483--492, 2017.

\bibitem{2017icra_zhu}
Yuke Zhu, Roozbeh Mottaghi, Eric Kolve, Joseph~J. Lim, Abhinav Gupta, Li
  Fei-Fei, , and Ali Farhadi.
\newblock Target-driven visual navigation in indoor scenes using deep
  reinforcement learning.
\newblock In {\em International Conference on Robotics and Automation (ICRA)},
  2017.

\end{thebibliography}
}

\begin{appendices}
~
\setcounter{table}{0}
\setcounter{figure}{0}
\renewcommand{\thetable}{A\arabic{table}}
\renewcommand{\thefigure}{F\arabic{figure}}

\begin{table*}[!htb]
    \begin{center}
    \resizebox{\linewidth}{!}{
    \begin{tabular}{lcccc}
    \toprule
Dataset & Number of Train Scenes & Number of Train Episodes & Number of Val Scenes & Number of Val Episodes \\
\midrule
\thirddataset{}   & 990                             & \phantom{0}898267                      & \phantom{0}905                                & 99                           \\
MP3D    & \phantom{0}61                              & 5000000                     & \phantom{0}495                               & 11                            \\
Gibson  & \phantom{0}72                              & 4932479                     & 1000                              & 16  \\
\bottomrule
    \end{tabular}
    }
    \end{center}
    \caption{\textbf{Dataset Statistics}}
    \label{table:dataset_stats}
\end{table*}

\FloatBarrier

\section{Dataset Details}
We constructed the Point-Nav datasets for each of \thirddataset{}, MP3D, and Gibson environments using a sampling-based method which filtered out easy episodes (those with $\frac{\text{euclidean distance}}{\text{geodesic distance}} < 1.1$). We additionally filter out episodes where there is no path between the start and goal location. The start points from these episodes were additionally used for the Exploration and Flee tasks, but the goal locations were ignored. Per-environment statistics are listen in Table~\ref{table:dataset_stats}.

\begin{figure}[t]
\begin{center}
\includegraphics[width=\columnwidth]{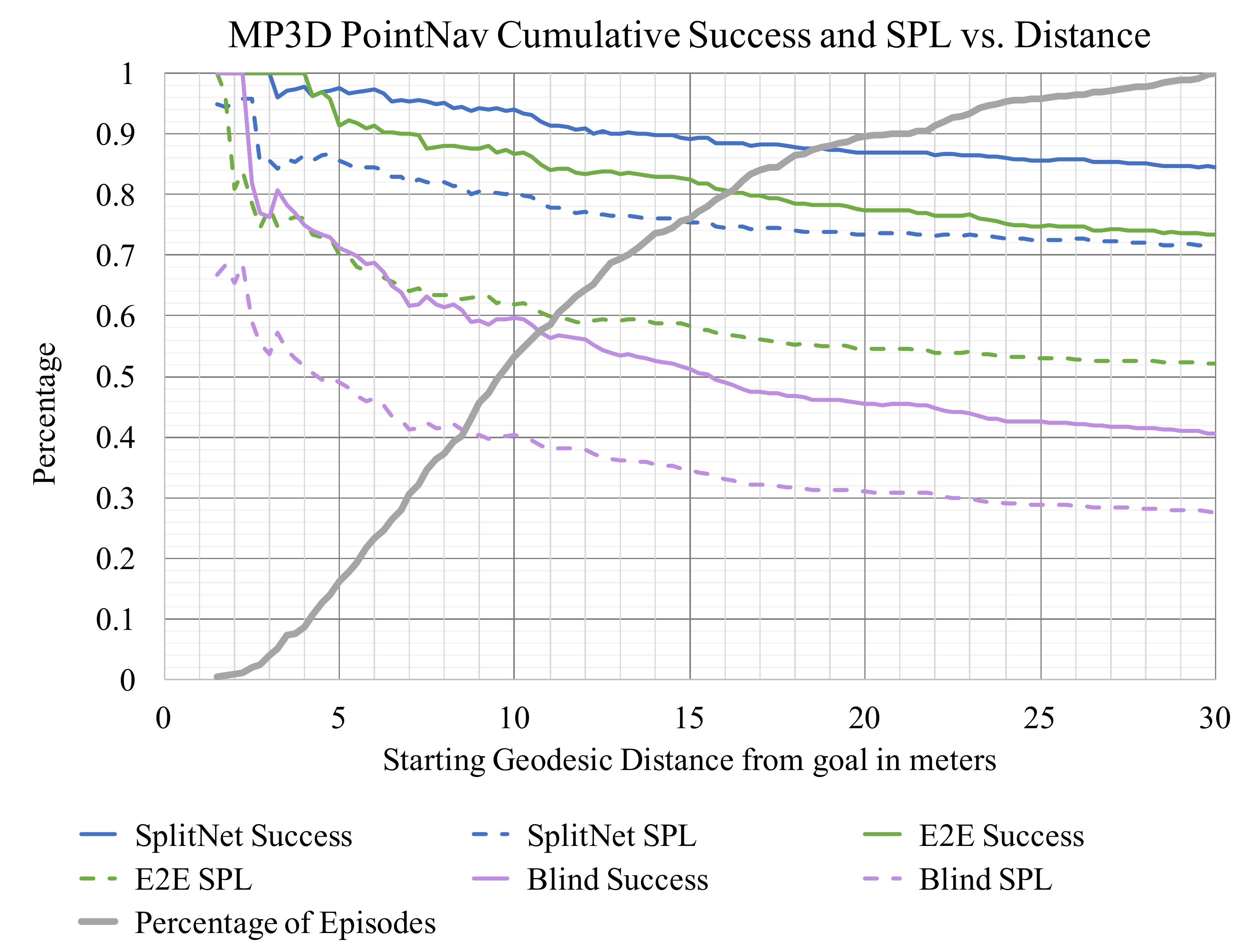}
\caption{Cumulative Accuracy and SPL on MP3D dataset.}
\vspace{-8mm}
\label{fig:cumulative}
\end{center}
\end{figure}
\section{Cumulative Performance based on Starting Distance}
To examine the balance of starting distances, Figure~\ref{fig:cumulative} shows a breakdown of performance on all episodes which start closer than N meters. This additionally shows that 50\% of the episodes start more than 9 meters away and nearly 25\% start greater than 15 meters away. 
\section{Network Architecture}
\begin{figure}[!htbp]
  \begin{center}
  \includegraphics[width=0.8\columnwidth]{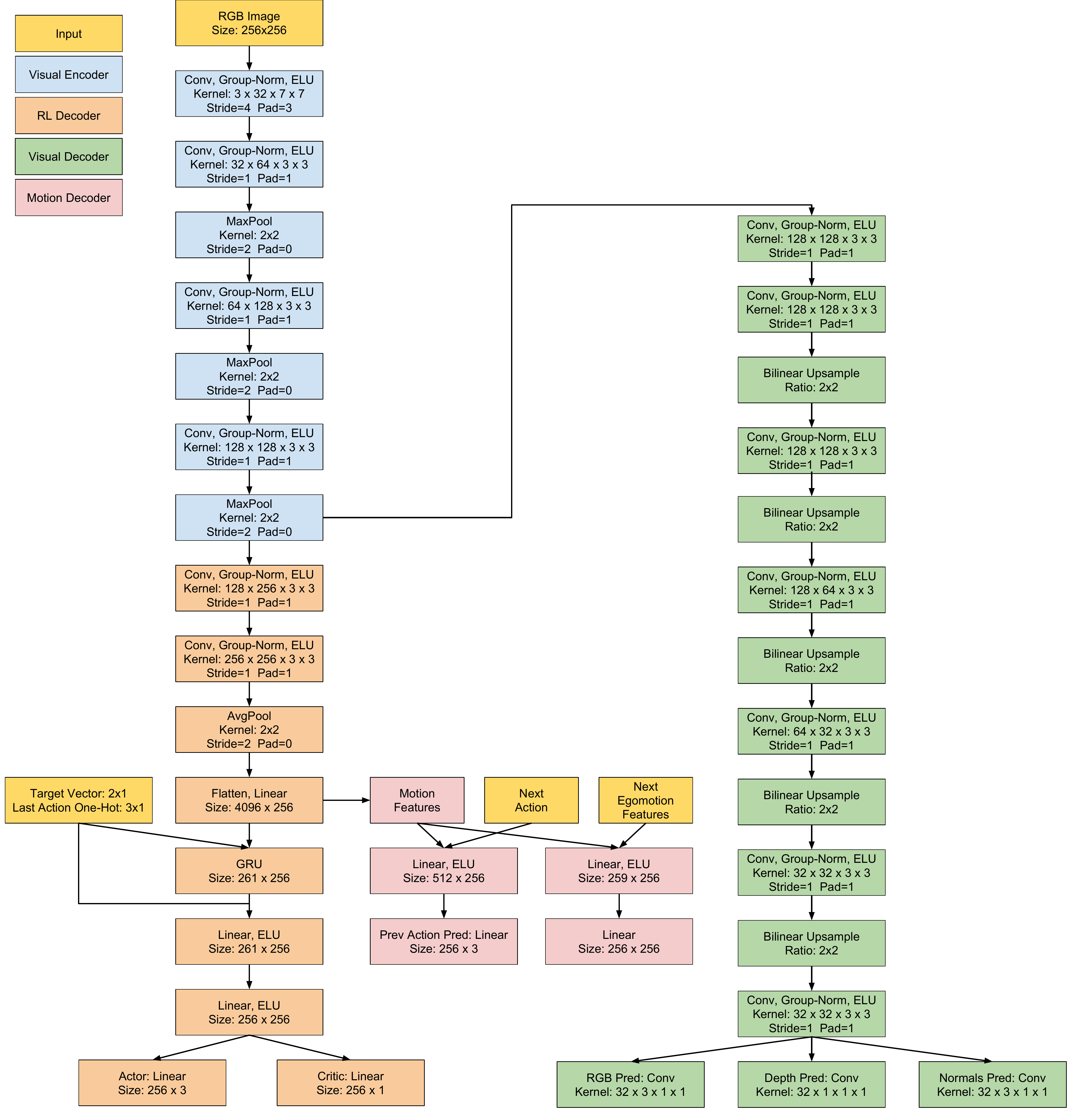}
  \end{center}
  \caption{\textbf{SplitNet Architecture}}
  \label{fig:network_arch}
\end{figure}

Figure~\ref{fig:network_arch} shows the encoder-decoder architecture of SplitNet. The E2E method trains the blue and orange portions, and the blind agent trains only the orange. Additionally the Motion layers are only trained for SplitNet. Those are omitted for simplicity due to them operating on multiple timesteps. The Egomotion 

\section{Auxiliary Outputs}
\begin{figure*}[t]
\begin{center}
\includegraphics[width=\linewidth]{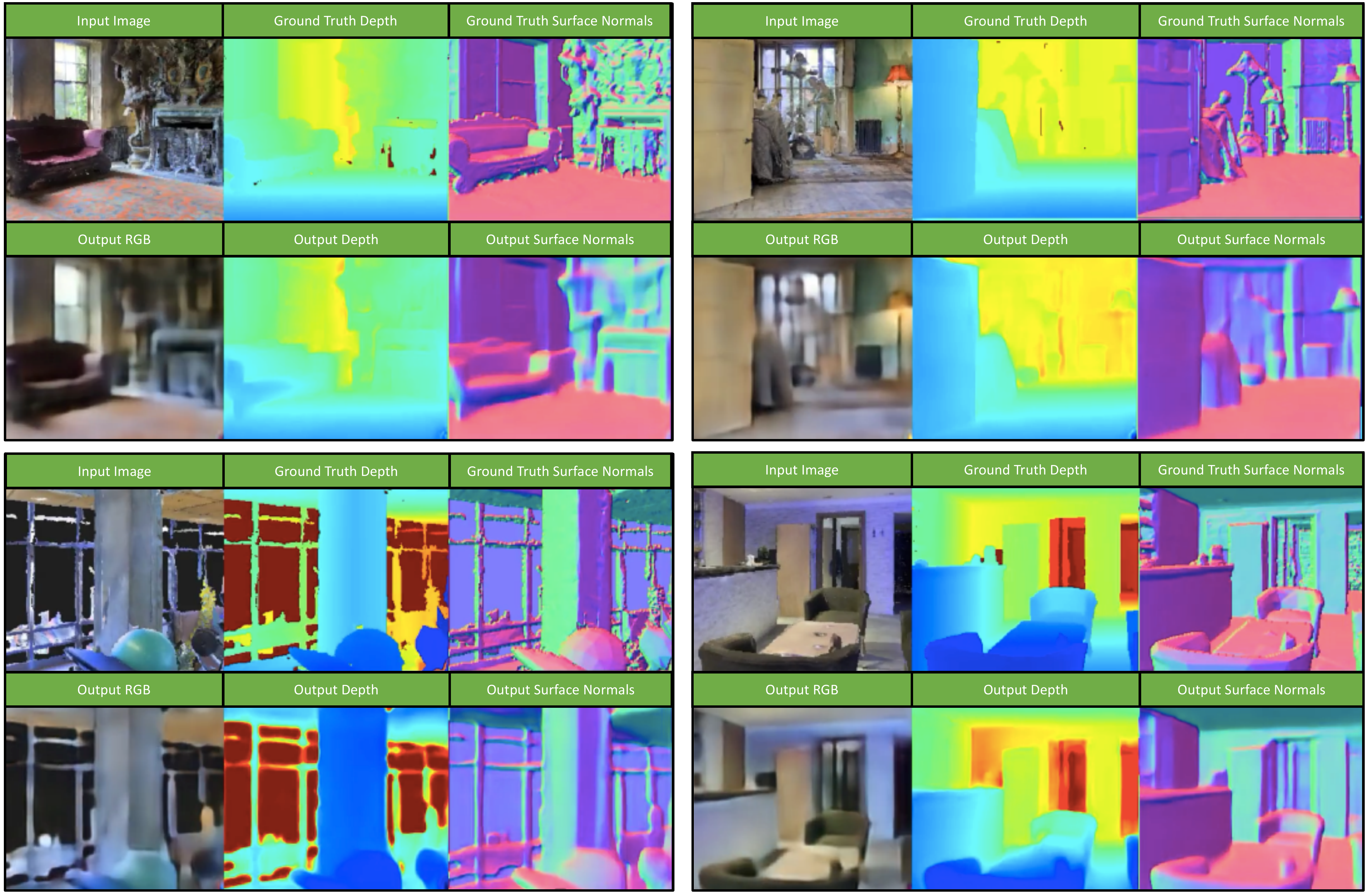}
\caption{Example predictions of auxiliary outputs on unseen MP3D test environments.}
\vspace{-8mm}
\label{fig:auxouts}
\end{center}
\end{figure*}

To verify that our network learns to encode geometry and appearance information, we show the output of the RGB, depth, and surface normal decoders on test environments in Figure~\ref{fig:auxouts}. Learning these encoder-decoders, especially for depth, improves the network's ability to codify visual information into actionable representations. The decoders also allow us to see where the network makes mistakes as a method of debugging the failure modes.

\vfill \break

\end{appendices}

\end{document}